\relax
\documentclass[letterpaper]{article} %DO NOT CHANGE THIS
\usepackage{aaai19}  %Required
\usepackage{times}  %Required
\usepackage{helvet}  %Required
\usepackage{courier}  %Required
\usepackage{url}  %Required
\usepackage{graphicx}  %Required
\usepackage{amsmath}
\usepackage{amssymb}
\usepackage{array}
\usepackage{bm}
\usepackage{algorithm}
\usepackage{algorithmic}
\usepackage{subcaption}
\usepackage{tabularx,ragged2e,booktabs,caption}
\begin{document}
\title{Nested multi-instance image classification}
\author{Alexander Stec\\ Engineering Sciences \\ and  Applied Mathematics \\ Northwestern University
\And Diego Klabjan\\ Industrial Engineering \\ and Management Sciences \\ Northwestern University  \And Jean Utke\\
Data, Discovery \& Decision Science \\ Allstate Insurance Company}
\maketitle
\begin{abstract}
There are classification tasks that take as inputs groups of images rather than single images.
In order to address such situations, we introduce a nested multi-instance deep network.
The approach is generic in that it is applicable to general data instances, not just images.
The network has several convolutional neural networks grouped together at different stages.
This primarily differs from other previous works in that we organize instances into relevant groups that are treated differently.
We also introduce a method to replace instances that are missing which successfully creates neutral input instances and consistently outperforms standard fill-in methods in real world use cases.
In addition, we propose a method for manual dropout when a whole group of instances is missing that allows us to use richer training data and obtain higher accuracy at the end of training.
With specific pretraining, we find that the model works to great effect on our real world and public datasets in comparison to baseline methods, with our improvements ranging from 1\% to 5\%.
\end{abstract}

\section{Introduction}\label{sec:introduction}
Deep learning, a relatively recent development in artificial intelligence, has achieved a great deal of success with many types of classification problems, ranging from speech to visual recognition. The latter has seen much success, and instrumental to this was the introduction of the convolutional neural network (CNN). When these networks are trained with large data sets and on high-performance computers, visual classification can achieve impressive results on large numbers of categories.

In the area of visual recognition, the majority of the research so far has related to singles instances of an image. Single instance is the case where each individual image, an instance, is tied to a label, and consequently there is no ambiguity of the label at the instance level. Perhaps the most well known data set of this type is ImageNet, which contains millions of labeled images. The single instance case is not comprehensive, however, and many real-world data sets do not have labels tied to individual instances, but rather to groups of instances. The groups of instances are commonly called bags, and these bags form the basis for training a classifier instead of individual instances.

The goals of multi-instance problems vary greatly depending on the data and desired application.  The earliest cases were just binary problems, but over the years this has expanded and there is no true ``typical" multi-instance case. In the context of visual recognition, a bag may be a large image and the instances are smaller crops of this image, or a bag could consist of several distinct whole images that are treated as instances. What type of training works best for multi-instance problems is highly problem specific. In some cases, the bag labels can be pushed to instance labels and the multi-instance problem is reduced to a single-instance problem. Often this is not appropriate and it is better to have a multi-instance network that takes all instances as input simultaneously.

\begin{figure}[ht]
\begin{center}
\includegraphics[width=8cm]{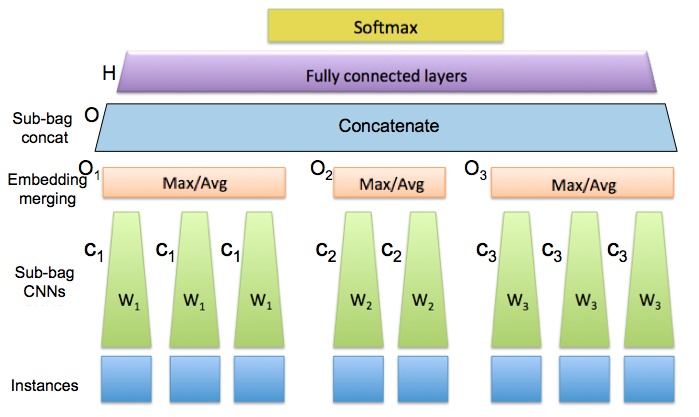}
\caption{Overview of the architecture for the nested multi-instance network.}\label{fig:revnet1}
\end{center}
\end{figure}

For the multiple instance network with images, it is common for copies of the network to share convolutional layers to get from the instances to embeddings. These embeddings can be merged in some manner and trained on in a standard way to classify at the bag level. In this paper, we introduce an expansion of the multi-instance model, which we call the nested multi-instance model. The nested multi-instance model groups instances within a bag to form sub-bags for every bag, and the overall architecture of the nested network can be seen in Figure \ref{fig:revnet1}. Our problem has instances that naturally fall into sub-bags, and so only these similar instances inside of a sub-bag  share convolutional layers to get their embeddings. After these embeddings are obtained, they are merged first at the sub-bag level and then at the bag level. At the sub-bag level, we use either an average or max over these embeddings to get the representation of the sub-bag, but we must concatenate at the bag level because the embeddings originate from different subspaces. This allows us to fine tune the weights shared within sub-bags to a higher degree, since the gradient is not affected by a vastly different instance.

To provide a practical application context for the method, consider a website for reselling used wedding dresses from user to user, for example preownedweddingdresses.com\footnote{https://www.preownedweddingdresses.com/used-wedding-dresses}. Each dress for sale has some combination of images and structured information, but these are not always complete. The different sub-bags for images correspond to different perspectives of the dress, such as front, back, side, collar, or sleeves. From these images, we can partially extract information about the dress. This includes, for example, filling in missing fields from the structured data, or evaluating other important aspects such as style, remaining value, and damage. Given the nature of the data, the sub-bags can have different numbers of images from dress to dress, and in some cases one of the sub-bags may not have any images at all. 
While this paper was originally inspired by a problem in the insurance space, it is applicable in more general cases.
The problem can be cast as a classification task based on multiple views of complicated objects where no single view can a priori be determined to be the most relevant, but rather a combination of views combined with an understanding about the typical types of views induces a grouping that can be exploited by specializing parts of the network.
The reported results are on a proprietary data set containing several million images and an appropriately constructed multi-view dataset based on Shapenet, a large-scale dataset of 3D shapes.
\begin{figure}[ht]
\begin{center}
\includegraphics[width=8cm]{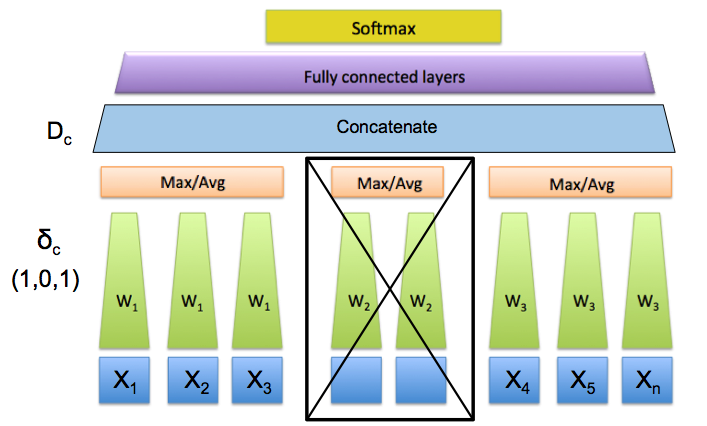}
\caption{Network with second sub-bag missing and one missing instance in the third sub-bag. The missing instance is filled in with a neutral instance $x_n$.}\label{fig:revnet2}
\end{center}
\end{figure}

This paper describes three major contributions. The first one is the introduction of the {\bf nested multi-instance model}. We have created a network which takes multiple sub-bags as input and then runs the instances within the same sub-bag through a shared CNN to get embeddings. These embeddings are merged at the sub-bag and bag level, before being used as input to a standard feed forward network. Second, we introduce {\bf methods to deal with missing instances within a sub-bag}.
Sub-bags must have set sizes and must be fully populated for training and testing, due to the requirement of deep learning frameworks we utilize to have a static computational graph.
There is software that allows for dynamic graphs, such as PyTorch and TensorFlow's most recent release, but it is unclear if this would actually be appropriate for our model since dynamic settings are much slower than static.
We combine the use of present instances and auxiliary optimization developed herein to populate the sub-bags in neutral manner. The final one is unique to our model, a {\bf method that addresses missing an entire sub-bag} from a bag. We address this using manual dropouts and an organized approach with regard to the order in which the given data is trained and the underlying minibatches.

\section{Related work}
The goals and methods used for previous multi-instance work vary greatly depending on the problem characteristics and available data. The goals can be placed into three broad categories, \cite{amores2013multiple}, called instance space, bag space, and embedded space paradigms. The instance space paradigm assumes the discriminative information is at the instance level and so creates classifiers at the instance level. This method then uses an aggregate of all the outputs to classify at the bag level \cite{xu2014deep,maron1998framework,andrews2003support,amores2013multiple}. The bag space paradigm assumes that the discriminative information is contained at a bag level. This means the bags are compared for similarities and differences, and these comparisons are then used as input for a standard learning algorithm \cite{wang2000solving,zhang2007local,belongie2002shape}. Our problem falls in the embedded space paradigm. In this paradigm, the whole bag is mapped to a single feature vector, and this feature vector can be used as input to standard learning algorithms \cite{opelt2006generic,serre2007robust,chen2004image,foulds2008learning,bunescu2007multiple,zhang2009multi}.

In prior work using images, the instances are often crops from larger images. For example, this is done with medical images. Here, the bag comprises the set of crops. In an application for detecting cancer in an image, the bags have binary (cancerous / non-cancerous) labels, and one cancerous instance (image crop) means the whole bag is cancerous. \cite{xu2014deep} use a shared instance level classifier for the image crops, and then put a softmax classifier over the instance level outputs to get the bag level prediction. This instance space work has previously been used in \cite{xu2012multiple}, where these crops give a pixel-wise classification of cancer for the whole image. 

\cite{hou2015patch} handle gigapixel-size, labeled images, and, similarly to the above, treat the whole image as a bag and the crops as the instances. They employ a two-stage training model. The first stage trains on crops with instance level classifiers to see which crops give the strongest responses. From these a pre-set number of crops are then chosen as the instances and trained by a multi-instance network. This is done in the embedding space paradigm, so the instances share a CNN and the outputs of the instance-level networks are then trained for the bag-level prediction. A similar two-stage embedded-space-paradigm training is seen in \cite{yan2016multi}. The authors aim to correctly identify body parts from images of internal body slices. They use crops instead of the whole image since many body parts look the same except for a few key regions. The first stage of training uses an instance-level classifier to try and find regions that get a high response, using max pooling, and the second stage uses these regions as instances and trains over the instance-level classifier to get the label for the bag. Other similar work includes \cite{zhu2016deep}, who use the same approach for mammogram classification, but expand on the max pooling idea described in \cite{yan2016multi} by combining instances.

With respect to image replacement optimization, related work has been done on inverting deep representations of images. \cite{mahendran2015understanding} found that it is possible to accurately reconstruct an image from its deep network embeddings using CNNs. The goal of  reconstruction is to generate an image that matches the original in the embedding space, but generally the solution is not unique. In order to reduce the solution space and get a more natural representation, they make use of natural image priors in their algorithm.  For our purposes we need only an image that acts as a neutral element in the context of merging the embeddings within sub-bags. Our work in this respect heavily relies on DeConv networks, as proposed by \cite{zeiler2014visualizing}. Finally, we rely on the concept of dropouts, proposed by \cite{srivastava2014dropout}, to deal with missing sub-bags.

An alternative way to handle image replacement, that is, a varying number of images per sub-bag, would be by means of recursive neural nets \cite{socher2011parsing,socher2011semi}. These group ``words" in a hierarchical way and within a group process words one by one thus allowing to have a flexible number of words per group. Such an approach is not adequate in our context since recursive neural nets require an order on words in a group and groups in a tree. In our case there is no order on images and imposing random orders is not justifiable.

\section{Nested multi-instance model}
\subsection{Model overview with non-variable bag composition}

Standard multi-instance learning is done by grouping the instances together in a bag\footnote{We intentionally use the less restrictive notion of bag in contrast to set in order to allow duplicates of instances in the (sub-) bags.}. Each instance in the bag is not labeled individually, but rather the bag is labeled as a whole. In our model we have multiple sub-bags $\cal S$ within a bag $B$, that is, a tuple of sub-bags forms a bag. Instances are grouped into sub-bags by specific characteristics. If we have $s$ sub-bags and $I_j$ instances in the $j$th sub-bag, we can write\begin{eqnarray*}
B &=& \left( {\cal S}_{1}, {\cal S}_{2},\ldots,{\cal S}_{s}\right),  \\
{\cal S}_{j} &=& \big\{ x_{j1}, x_{j2},\ldots, x_{jI_j}\big\}, 
\end{eqnarray*}
where $x_{jk}$ is the $k$th element in $j$th sub-bag ${\cal S}_j$. In addition, each bag is associated with a label $y$. If there are $M$ classes, then each $y$ can take one of $M$ values. Each training example can then be thought of as $(B, y)$, the pair of the bag and its associated label. The whole training set is denoted
\begin{equation}\label{eqn:trainset}
\mathcal{B} = \big\{ (B^{1}, y^{1}), (B^{2}, y^{2}), \dots, (B^{K}, y^{K}) \big\}
\end{equation}
with $K$ bags and associated labels.
\\ \\
We start by describing how one bag, $B$, is processed. Each sub-bag ${\cal S}_j$ is associated with a distinct convolutional network $C_j$, so we have $s$ convolutional networks. At this first stage, for every sub-bag ${\cal S}_j$ each instance is fed forward through the corresponding network $C_j$. This can be represented as
\begin{eqnarray*}
C &=& \left( \mathcal{C}_{1}, \mathcal{C}_{2}, \dots, \mathcal{C}_{s}\right), \\
\hspace{.2cm} \mathcal{C}_{j} &=& \big\{ C_{j}\left( x_{j1}\right), C_{j}\left( x_{j2}\right), \dots, C_{j}\left( x_{jI_j}\right)\big\}, 
\end{eqnarray*}
where $C$ is an $s$-tuple whose elements are bags with cardinality $I_j$, see Figure \ref{fig:revnet1}. Each element in these sets is the output of an instance being run through the convolutional network. Thus each $C_{j}\left( x_{jk}\right)$ is a vector flattened from a matrix that represents two-dimensional convolutions of an image.
\\ \\
Now we introduce the aggregation function $f_{a}$ to reduce each set of output vectors to a single output vector. For $f_{a}$ we explored element-wise average and max.
Applying the aggregation function, we get the output tuple 
\begin{eqnarray*}
\left(\bm{O}_1,\bm{O}_2,\ldots,\bm{O}_s\right)=\left( f_a(\mathcal{C}_1), f_a(\mathcal{C}_2),\ldots, f_a(\mathcal{C}_{s})\right).  
\end{eqnarray*}
The last step is to combine the sub-bag embeddings by concatenating these vectors into one vector denoted as  
\begin{eqnarray*}
\bm{O} = f_a(\mathcal{C}_1)\diamond f_a(\mathcal{C}_2)\diamond\ldots\diamond f_a(\mathcal{C}_{s}). 
\end{eqnarray*}
This vector is then used as the input for a standard fully connected network. The fully connected layers produce a vector $\bm{H} = H\left( \bm{O} \right)$, where $\bm{H}\in \mathbb{R}^M$, with $M$ being  the number of classes. The softmax function is applied to $\bm{H}$, yielding a vector of probabilities for each class, $\bm{P}\left(B\right) \in \mathbb{R}^M$. The loss $L$ is calculated over all $K$ bag and label pairs, and is written as
\begin{equation}\label{eqn:loss}
L = \sum_{i=1}^{K}D_{K\!L}\big(y^{i}\,||\,\bm{P}(B^{i})\big) , 
\end{equation}
with $D_{K\!L}$ representing the Kullback-Leibler divergence.
 
All of the above assumes both that all sub-bags are of the same size and all sub-bags are present for each bag. Given the nature of the data, these assumptions do not hold. The following two sections describe the ways in which we account for variable sub-bag size and missing sub-bags.

\subsection{Variable instances within a sub-bag}\label{sec:varInstances}
Deep learning packages, such as Tensorflow and Theano, require a fixed computational graph implying a static network topology. For this reason, it is necessary for the network to have a fixed size. This means that the number of sub-bags and number of instances per sub-bag must be fixed for each bag. In this section we describe our approach to accommodate a variable number of instances per sub-bag while still assuming that all sub-bags are present for each bag.

Each sub-bag ${\cal S}_j$ has an associated set of images from which we create the required number of instances $I_j$. In the simplest case, we create one instance per image when the size of the image set equals $I_j$. We must be able to handle bags that contain one or more sub-bags with different numbers of images. No matter how many sub-bags need to be ``modified," each one can be done independently, and so we discuss how to handle just one sub-bag. Let $m$ be the number of images in the set and let $I$ be the sub-bag size and $C$ be the corresponding network. 

If $m > I$, we randomly select $I$ distinct images during each epoch of training and at test time. Once the $I$ images are chosen, an instance is created from each image by a combination of scaling and taking a random crop or by taking a center crop in order to match the required size of the CNN. For the final classification using the trained network one would employ a similar approach with random samples of the images and choose as final result among the vectors $\bm{P}$ yielded by the samples that vector with the highest maximum norm or a similar selection criterion.  

If $ m < I$, we propose three methods to generate the $n = I - m$ instances from $I$ images.
\\ \\
{\bf Reproduction:}
With {\bf max exact fill} the sub-bag is first filled as much as possible by replicating images. 
This is done by replicating the $m$ images $\lfloor I / m \rfloor$ times. 
Then, for each image, an instance is created by taking a center crop of the image. 
Both the max and average aggregations are invariant to this approach.  
This may leave $o=I\bmod m$ open instances in the sub-bag. With {\bf random reproduction} these remaining instances are chosen by randomly selecting from the same number of images and taking a different crop from each of these images. 
\\ \\ 
{\bf Random fill:} 
Instances are created by taking random center or corner crops of images randomly selected from the $m$ images without replacement. 
Once the set has been exhausted, the random selection starts again with the full image set until all $I$ instances are created.
\\ \\
{\bf Optimization:} 
We first fill in the sub-bag with reproduction max exact fill. 
Under the max aggregation, the remaining $o$ images can be replicated individually again to fill the  sub-bag, because the max aggregation remains invariant to this fill. 
For average aggregation, however, we first calculate the element-wise average 
\begin{equation*}
\mu=\frac{1}{I-o}\sum_{i=1}^{I-o}C(x_i)
\end{equation*}
of the embeddings obtained for the $I-o$ instances we filled so far. 
We create a neutral instance $x_n$ for which $\mu=C(x_n)$, that is, the average aggregation is invariant to filling the remaining $o$ slots with $x_n$.
The neutral instance $x_n$ is created by solving an auxiliary optimization problem $\mu$. 
This makes use of deconvolutional layers that reverse the sub-bag's convolutional network. 
In this step, the network $C$ itself does not change. 
Specifically, we are solving
\begin{eqnarray}\label{eqn:neutralOpt}
x_n=\underset{x\in X}{\mbox{argmin}} \| C(x) - \mu \|^{2}_{2} 
\end{eqnarray}
by stochastic gradient descent where $X$ is the space of arbitrary instances suitable as input to $C$. 
In order to easily compute the gradient in the given CNN implementation framework, we add an extra layer corresponding to the input (the original input layer now becomes the first bottom layer). This new input layer has a neuron for each pixel and channel. Each input neuron is connected to one and only one neuron in the first bottom layer (the original input layer). The weights on these connections are free and correspond to $x$. The first bottom layer is then followed by network $C$ with all weights fixed. It is straightforward to see that this network models (\ref{eqn:neutralOpt}) and thus standard backpropagation with respect to $x$ can be applied.
% It is trained only with examples consisting of an all-one vector. {\color{red}[JU: I am not sure how to relate this last sentence to the above without further explanation]}

%We start the optimization algorithm by running $x_{t}$ through the deconvolutional layers to get an initial image. This problem is solved by the stochastic gradient. The only trainable layer is a one-to-one pixel mapping layer that takes the initial image to the image run through the CNN. The gradient is the same as that for $\left( 2 \right)$.

\subsection{Sub-bag dropout} \label{sec:dropout}

In standard dropout, random units are dropped out of the network with a uniform probability at each training instance. For our model, we are interested in dropout at the sub-bag level applied to the embeddings, $\bm{O}$, that are the input of the fully connected network $H$. Consequently, no individual units, but rather whole sets of units $\bm{O}_{j}$ are dropped. With $s$ sub-bags as candidates for dropout there are $2^{s}$ possible configurations. To represent each configuration $c$ we introduce a tuple, $\delta_c \in \{0,1\}^s$,that is, a tuple of $0$s and $1$s of length $s$. A $1$ corresponds to the sub-bag being present in $c$ and a $0$ corresponds to a missing sub-bag in $c$. Figure \ref{fig:revnet2} shows a case of one missing sub-bag. 
%{\color{red}[JU: fix up figures 1 and 2 to match notation]}

Before passing $\bm{O}$ as input to the fully connected layer, the sub-bag dropout is performed by\begin{eqnarray*}
\bm{d}_{jc} &=& \bm{1}_{j} \cdotp \delta_{cj},\quad j=1,\ldots,s  \\
\bm{d}_{c} &=& \bm{d}_{1c} \diamond \bm{d}_{2c} \diamond \dots \diamond \bm{d}_{sc}  \\
\bm{D}_{c} &=& \bm{d}_{c} \Delta \bm{O} 
\end{eqnarray*}
where the length of $\bm{1}_{j}$ and $\bm{O}_{j}$ are the same and equal to $I_j$ and $\bm{D}_{c}$ becomes the input to the fully connected network. Note that $\bm{d}_{jc}$ is either a $0$-vector or a $1$-vector. Here, $\Delta$ is the Hadamard product. The output of the fully connected layers is then $\bm{H}_{c} = H\left( \bm{D}_{c} \right)$. Again, the softmax function is applied to the output vector to get a vector of probabilities $\bm{P}\left( B_{c}^{i} \right)$ calculated over a chosen configuration. The loss function (\ref{eqn:loss}) becomes
\begin{eqnarray}
L = \sum_{c=1}^{2^{s}} \sum_{i \in {\cal K}_{c}} D_{K\!L}\left( y^{i}\middle\|{\bf P}\left(B_{c}^{i}\right) \right), \nonumber
\end{eqnarray}
where the training set ${\cal B}$ and the corresponding set $\cal K$ of $K$ indices  are partitioned such that $i \in {\cal K}_{c}$ iff $B^i$ has at least one instance for each of the sub-bags present in configuration $c$.

Using standard dropout, at test time all the nodes would be made active and the weights would be scaled. However, since our dropouts are constructed with our data in mind, even the test bags will need to have dropout applied.
%A diagram of the network connectivity can be seen in Figure \ref{fig:revnet1}.
%\begin{figure}[htbp]
%\begin{center}
%\includegraphics[width=6.25in]{dropout}
%\caption{Illustration of the connectivity in a fully connected network, a standard dropout network, and our sub-bag dropout network.}
%\end{center}
%\end{figure}

\section{Training}
Training the whole network from scratch over all configurations is a very hard task. Therefore, pretraining is critical for learning. We adopt VGG16, introduced in \cite{simonyan2014very}, as the template for the CNN networks $C_i$ but use only the convolutional layers within our large nested multi-instance network as opposed to the VGG16 network with all layers which we will denote as $\bar{C}_i$. Even when sharing convolutional networks within sub-bags, starting from random weight initialization is intractable.

\subsection{Pretraining} \label{sec:pretrain}
We bootstrap our approach by using VGG16 weights trained on ImageNet data. ImageNet is sufficiently general to train filters that are useful on our given data, however, this applies more to the lower layers than to the final layers. To start pretraining, we pair bag labels $y^j$, see (\ref{eqn:trainset}), with the individual images in $B^j$ and train each sub-bag network $\bar{C}_i$, in standard single instance fashion on the images in $B^j$ that pertain to the $i$th sub-bag. At the end of training, we have $s$ separately trained networks. We then cutoff the fully connected part of the network $\bar{C}_i$, and save only the weights for the convolutional layers for $C_i$. The specific phases of the training are discussed next.

\subsection{Training phases}
We begin training over bags using the pretraining weights for all $s$ sub-bag networks $C_i$ as initializations for the convolutional weights of the full network and random initialization for the fully connected layers. Throughout all training phases exponential decay is used to control the learning rate. In the previous section we already mentioned the configuration-induced partition of the training set, ${\cal B}=\bigcup\limits_{c=1}^{2^s}{\cal B}_c$.

{\bf Phase 1:} To boost convergence, we start with the full configuration $\hat{c}$,  $\|\delta_{\hat{c}}\|_1=s$, that is, all bags in ${\cal B}_{\hat{c}}$ have at least one image for each sub-bag.  Consequently, there is no need for sub-bag dropouts during this part of the training, but the missing images within sub-bags still are accounted for as outlined in the previous section. Minibatches are created from the elements in ${\cal B}_{\hat{c}}$; all layers are unlocked. 

{\bf Phase 2:} The weights updated during Phase 1 serve as initialization for Phase 2. This phase has an outer sequence governing weights locking as follows.
\begin{description}
\item[FC1:] unlock weights in $H$, i.e. the fully connected layers, and lock all weights in the $C_i$, i.e. the convolutional layers
\item[CL:] lock all weights in $H$, unlock convolutional layers in the $C_i$ one at a time from top to bottom
\item[FC2:] unlock weights in $H$, lock all weights in the $C_i$
\end{description}

For each of the steps in the sequence we train on all elements in $\cal B$ but reset the learning rate before each step. 
Furthermore,  because the dropout configurations differ among the subsets ${\cal B}_c$ and the dropout is effected by modifying the full network definition, we group  minibatches by the ${\cal B}_c$ they were created from. 
By the same argument stipulating Phase 1, we partially order the ${\cal B}_c$ and their associated minibatches for the epoch by starting with the richest information, i.e. ${\cal B}_{\hat{c}}$, where $||\delta_{\hat{c}}||_1=s$ and then by decreasing values yielded by  $||\delta_c||_1$ attaining $s-1,\ldots,1$ and random order in case of ties.

\section{Experiments}
\textbf{Real world datasets:}
For the experiments described in this part we used a real world proprietary data set and derived training sets with classification labels originating in the insurance space.
While we cannot share qualitative details about the data, we can disseminate quantitative aspects to illustrate accuracy results.
We report on two distinct classification tasks denoted as ``Case 1'' and ``Case 2.''
In each bag, we have three sub-bags, but the group of instances represented by each sub-bag differs between the two cases.
The test set ${\cal B}_T\subset{\cal B}$ consists of $10\%$ of randomly chosen bags in all experiments.

For Case 1, we have more than 500,000 bags with 3 sub-bags, and of these approximately $60\%$ are full configuration bags.
Each bag is assigned one of four possible labels.
For training Phase 1, the layers in the $C_i$ are initialized using the weights resulting from pretraining, 
see the Training section, while the weights in 
$H$ are chosen by Glorot uniform initialization.
%After Phase 1, we test these weights on all configurations to get a baseline of the accuracy for the full set of bags before starting Phase 2.

Table \ref{tab:case1} contains validation accuracies for various stages of training with the different approaches to filling in missing instances discussed previously.
In addition to this, the table also contains baseline results to show the increased performance of the proposed network.
The first of these baseline results, listed under column ``Shared sub-bags,'' uses shared weights amongst the different sub-bags and the optimization fill-in method.
The second (``Individual nets'') takes the scores from the pretrained individual networks for all of the present instances, and then averages these scores to make a prediction for the bag.
%Of course, the highest accuracies are obtained when we train only on full configurations.
%This intuitively makes sense since these configurations have the most complete information.
%When switching to training over all configurations, the accuracy drops because of the inclusion of bags that have not yet been trained on and that have missing sub-bags.

Training Phase 2 is able to significantly improve the accuracy over all configurations, with excellent improvement in both the first fully connected training stage (FC1) and in the convolutional layer unlocking stage (CL).
The table clearly shows that training the entire model is beneficial and that the naive approach of considering only full configurations is fairly weak.
The weights obtained at the end of Phase 1 applied only to full configurations ${\cal B}_T\cap {\cal B}_{\hat{c}}$ yield an accuracy of $93\%$, which is, as expected, much higher than the accuracy of $68\%$ of the same weights tested on all configurations.
This drop is expected since at that point the model has not seen any of the bags with missing sub-bags.
Table \ref{tab:case1} also exhibits the comparison of different image fill-in strategies. While the difference is not big, our new optimization method consistently outperforms the remaining two strategies.
Further, the method with shared weights performs worse at all stages of training, justifying the varying sub-bag weights.
Worse still is the aggregate prediction of the individual instances, indicating that the multiple views are synergistic and should not be aggregated naively.

\begin{table}[htp]
\begin{center}
{\small
\resizebox{\columnwidth}{!}{
\begin{tabular}{|l|c|c|c|c|c|}
\hline
 & Reproduction & Random fill & Optimization & Shared sub-bags & Individual nets  \\
%\hline
%Phase 1 on ${\cal B}_{\hat{c}}$ &  92.7 & 92.9 & 93.3\\
\hline
Phase 1 &  68.4 & 68.5 & \bf{68.8} & 67.9 & -\\
\hline
Phase 2, FC1 & 72.2 & 72.3 & \bf{72.5} & 71.5 & -\\
\hline
Phase 2, CL & 75.0 & 75.1 & \bf{75.3} & 73.3 & -\\
\hline
Phase 2, FC2 & 75.6 & 75.7 & \bf{75.9} & 74.1 & 67.5\\
\hline
\end{tabular}
}
}
\end{center}
\caption{Test accuracies in percent for Case 1 with 3 instances per sub-bag.}\label{tab:case1}
\end{table}%

For Case 2, we have about 100,000 bags with 3 sub-bags, and of these approximately $40\%$ are full configuration bags.
Each bag is assigned one of seven possible labels, and the experiments performed are the same as in the first case.
In Table \ref{tab:case2}, the same behavior is seen in the switch from full configurations to all configurations, with the accuracy on the full configuration test set at the end of Phase 1 being $84\%$.
Phase 1 again achieves the highest accuracies, with a large drop when tested on all configurations.
For this case there is good improvement when training over the fully connected layers in Phase 2, but training over the convolutional layers does not significantly boost accuracy.
It is possible that the convolutional filters initialized with the pretraining weights and further tuned in training Phase 1 were already sufficient over all configurations, and consequently only little was to be gained by training these layers further.
%A summary of the results over the different methods to fill in missing instances is found in Table 2.

\begin{table}[htp]
\begin{center}
%\begin{adjustbox}{width=\columnwidth-10pt}
\resizebox{\columnwidth}{!}{
\begin{tabular}{|l|c|c|c|c|c|}
\hline
& Reproduction & Random fill & Optimization & Shared sub-bags & Individual nets  \\
\hline
%Phase 1 on ${\cal B}_{\hat{c}}$ &  84.2 & 84.4 & 84.5\\
%\hline
Phase 1 &  64.1 & \bf{64.2} & \bf{64.2} & 63.6 & -\\
\hline
Phase 2, FC1& 68.2 & 68.3 & \bf{68.4} & 67.7 & -\\
\hline
Phase 2, CL & 68.3 & 68.5 & \bf{68.6} & 67.7 & -\\
\hline
Phase 2, FC2 & 68.4 & 68.6 & \bf{68.7} & 67.9 & 62.9\\
\hline
\end{tabular}
}
%\end{adjustbox}
\end{center}
\caption{Test accuracies in percent for Case 2 with 3 instances per sub-bag.}
\label{tab:case2}
\end{table}%

\textbf{Shapenet dataset:}
We also use a dataset created from the Shapenet 3D dataset \cite{chang2015shapenet}.
The artificial dataset was created by taking 2 images of 3D objects from each of 3 different perspectives (top, front, back).
One of the two images is taken directly from the perspective of its respective sub-bag (e.g. a direct front view) and the other is taken at an angle offset from the first image by 30 degrees.
This is done so that the two images composing each sub-bag are not too similar, i.e. from nearly identical perspectives, and also so that the offset image does not stray into the perspective of another sub-bag.
As with the real world data sets, the test set ${\cal B}_T\subset{\cal B}$ consists of $10\%$ of randomly chosen bags.

For this dataset, we have approximately 40,000 bags with 3 sub-bags corresponding to top, front, and back views of the objects.
This dataset has 13 classes after filtering out objects that have very low representation (less than 1 percent of the 40,000).
This dataset has no natural missing instances or missing sub-bags, thus we explore different methods of dropping instances and bags to report the model performance under varying circumstances.
We consider three methods to create missing instances.
For each of these three methods, anywhere from 0 to 4 instances are removed with uniform probability.
The difference in the methods is in how these instances are chosen once we have determined how many to remove.

The first method is the simplest, and consists of randomly dropping any instance with a uniform probability.
The next two methods deal with the relevance of the instances.
We measure instance relevance by looking at the correct class probability for each instance using the individual pretrained networks.
One method drops the most relevant instances (those with the highest class probability), while the other drops the least relevant instances.
Dropping instances naturally leads to missing sub-bags as well, so it was not necessary to further drop sub-bags after instance dropping.
After random instance dropping we are left with $80\%$ full configuration bags, after most relevant instance dropping  $45\%$, and after least relevant dropping  $48\%$.
%The generated instance can be found at \url{https://www.github.com/alex-stec/shapenet-multi-view}.

The same experiments are performed as in the real world cases.
We report the results for each of the three instance dropping methods in Table \ref{tab:shapenet}.
The accuracies on the full configuration test sets at the end of Phase 1 are $95\%$, $91\%$, and $93\%$ for random dropping, most relevant dropping, and least relevant dropping respectively.
It is important to note, however, that since each sub-bag has only 2 instances, the reproduction method always leaves the max or average aggregation unchanged since the same image is replicated.
Thus for this dataset, we observe how closely optimization performs relative to exact reproduction, where reproduction in this case is equivalent to the optimization method performing perfectly.

\begin{table}[htp]
\begin{center}
%\begin{adjustbox}{width=\columnwidth-10pt}
\resizebox{\columnwidth}{!}{
\begin{tabular}{|l|c|c|c|c|}
\hline
Random & Reproduction & Optimization & Shared sub-bags & Individual nets  \\
\hline
%Phase 1 on ${\cal B}_{\hat{c}}$ &  84.2 & 84.4 & 84.5\\
%\hline
Phase 1 &  \bf{90.3} & 90.1 & 89.5  & -\\
\hline
Phase 2, FC1& \bf{91.3} & 91.1 & 90.4  & -\\
\hline
Phase 2, CL & \bf{91.5} & 91.1 & 90.7  & -\\
\hline
Phase 2, FC2 & \bf{91.8} & 91.5 & 90.9 & 90.2\\
\hline
\end{tabular}
}
%\end{adjustbox}
\end{center}
\begin{center}
%\begin{adjustbox}{width=\columnwidth-10pt}
\resizebox{\columnwidth}{!}{
\begin{tabular}{|l|c|c|c|c|}
\hline
Most relevant & Reproduction & Optimization & Shared sub-bags & Individual nets  \\
\hline
%Phase 1 on ${\cal B}_{\hat{c}}$ &  84.2 & 84.4 & 84.5\\
%\hline
Phase 1 &  \bf{81.3} & 81.2 & 80.7  & -\\
\hline
Phase 2, FC1& \bf{83.0} & 82.9 & 82.1  & -\\
\hline
Phase 2, CL & \bf{83.5} & \bf{83.5} & 82.5  & -\\
\hline
Phase 2, FC2 & \bf{84.6} & 84.5 & 83.4 & 80.2\\
\hline
\end{tabular}
}
%\end{adjustbox}
\end{center}
\begin{center}
%\begin{adjustbox}{width=\columnwidth-10pt}
\resizebox{\columnwidth}{!}{
\begin{tabular}{|l|c|c|c|c|}
\hline
Least relevant & Reproduction & Optimization & Shared sub-bags & Individual nets  \\
\hline
%Phase 1 on ${\cal B}_{\hat{c}}$ &  84.2 & 84.4 & 84.5\\
%\hline
Phase 1 &  \bf{85.1} & 84.9 & 84.0  & -\\
\hline
Phase 2, FC1& \bf{86.2} & 86.0 & 84.9  & -\\
\hline
Phase 2, CL & \bf{87.1} & 87.0 & 85.7  & -\\
\hline
Phase 2, FC2 & \bf{87.5} & 87.4 & 86.8 & 85.3\\
\hline
\end{tabular}
}
%\end{adjustbox}
\end{center}
\caption{Test accuracies in percent for Shapenet dataset with 2 instances per sub-bag for each of the three instance dropping methods.}
\label{tab:shapenet}
\end{table}%

The behavior for each of the 3 methods is in general the same as the one observed in the real world datasets.
The random dropping method is similar to Case 2 in that there is not much improvement within Phase 2 of training, while the other two methods are more similar to Case 1.
It is not surprising that dropping the most relevant instances gives the lowest accuracy, but it is interesting that randomly dropping instances performs better than dropping the least relevant instances.
This is likely because the latter method leaves more empty sub-bags, and so the information from the dropped perspectives does not enter the model at all.
While those instances may be the least relevant on their own, it is seems as those they have an important effect in the full model that combines all the perspectives.
Finally, we observe that optimization closely tracks the reproduction method, meaning it is indeed performing its intended function.

\textbf{Additional Analysis:}
The previous results all use the average aggregation, but taking the max aggregation over the embeddings avoids the need for optimization since the aggregation is invariant to the reproduction of any number of instances in the sub-bag.
Table \ref{tab:maxOrAverage} shows a comparison for the best results obtained from a comparison between the max and average.
For this comparison, we use Case 2, optimization for the average aggregation, and random fill for max aggregation.
As stated previously, such a comparison is not possible on the Shapenet dataset with 2 instances per sub-bag, because any sub-bag with with one instance simply reproduces the existing instance.

\begin{table}[htp]
\begin{center}
%\begin{adjustbox}{width=\columnwidth-10pt}
%\resizebox{\columnwidth}{!}{
\resizebox{.62\columnwidth}{!}{
\begin{tabular}{|l|c|c|}
\hline
  & Max & Average  \\
\hline
Phase 1 on ${\cal B}_T\cap{\cal B}_{\hat{c}}$ &  84.1 & 84.5\\
\hline
Phase 1 on ${\cal B}_T$ &  63.7 & 64.2 \\
\hline
Phase 2, FC1 on ${\cal B}_T$ & 68.0  & 68.4 \\
\hline
Phase 2, CL on ${\cal B}_T$  & 68.3  & 68.6 \\
\hline
Phase 2, FC2  on ${\cal B}_T$ & 68.3 & 68.7 \\
\hline
\end{tabular}
}
%\end{adjustbox}
\end{center}
\caption{Comparison of max and average aggregations.}
\label{tab:maxOrAverage}
\end{table}%

In addition to testing methods for instance embedding aggregation, we also varied the maximum number of instances used in each sub-bag from 2 to 4.
In the case where 2 sub-bags are used, the optimization method is actually not required at all, since the reproduction fills the sub-bag whenever there is a missing instance.
Table \ref{tab:varyInstance} shows a summary of the results obtained for both Case 1 and 2 using the optimization method.
There is an advantage to using 3 instances over 2, but the difference between 3 and 4 seems not to be significant.
This last result is likely because only a very small percentage of sub-bags contain 4 or more instances in our data, and so further increasing the number of instances does not supply more complete information.
Rather, for most sub-bags, instances are just reproduced more, which increases the training time without the benefit of an accuracy increase.
We conclude that the optimal number of instances per sub-bag should be determined by the nature of the data, in particular the distribution of the present number of instances for the corresponding sub-bag.

\begin{table}[htp]
\caption{Instance and sub-bag analysis.}
\fontsize{10}{12}\selectfont
\begin{subtable}[t]{.45\columnwidth}
\caption{Test accuracy results for varying number of instances per sub-bag after training Phase 2.}
\label{tab:varyInstance}
\raggedright
\resizebox{\columnwidth}{!}{
\begin{tabular}{|c|c|c|c|}
\cline{1-4}
Inst. per bag & 2 & 3 & 4  \\
\cline{1-4}
Case 1 & 75.7 & 75.9 & 75.9\\
\cline{1-4}
Case 2 & 68.6 & 68.7 & 68.6 \\
\cline{1-4}
\end{tabular}
}
\end{subtable}
%\end{adjustbox}
\quad
\begin{subtable}[t]{.45\columnwidth}
\caption{Accuracy drops from applying sub-bag dropout to full configurations.}
\label{tab:dropout}
\raggedleft
\resizebox{\columnwidth}{!}{
\begin{tabular}{|l|c|c|}
\cline{1-3}
 Acc. Drop & Case 2 & Shapenet  \\
\cline{1-3}
Sub-bag 1 &  7.1 &  4.9\\
\cline{1-3}
Sub-bag 2  &  8.9 &  9.2\\
\cline{1-3}
Sub-bag 3 &  4.5 & 2.5 \\
\cline{1-3}
\end{tabular}
}
\end{subtable}
\end{table}%

\begin{figure}[ht]
\begin{center}
\includegraphics[width=6cm]{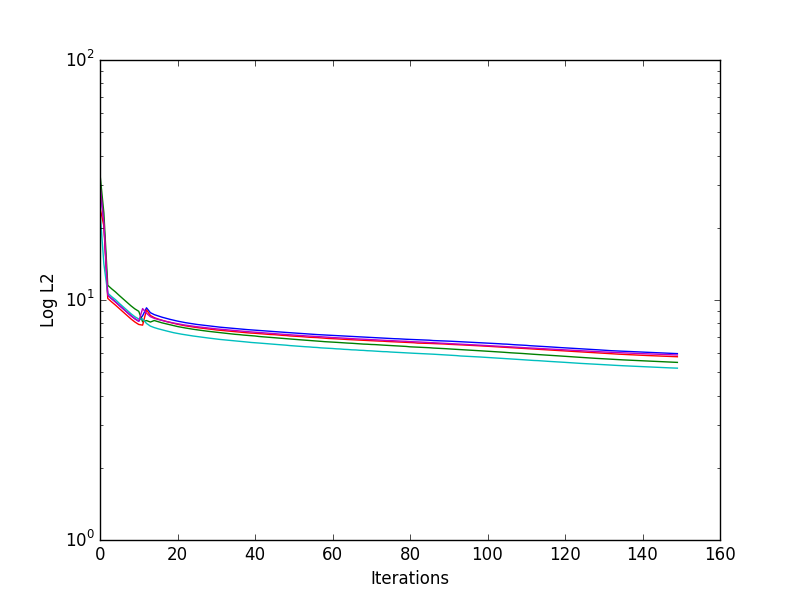}
\caption{$L_2$ norm versus iterations for auxiliary optimization.} \label{fig:l2norm}
\end{center}
\end{figure}

As seen in Table \ref{tab:case1} and Table \ref{tab:case2}, the optimization method was able to achieve the highest accuracies out of all the methods.
Further, as Table \ref{tab:shapenet} shows, the optimization method also closely tracks exact reproduction in the case of two instances, where the neutral instance is the one that gives the same embedding as the instance which is present.
This difference is more pronounced in Case 1 than in Case 2, and this is likely due to the nature of the instances rather than a varying effectiveness of the optimization approach.
%This suggests that the effectiveness of using optimization is dependent on the type of instances for that case.
%For some cases, the random cropping and/or the reproduction method will be sufficient to provide similar accuracy results with the benefit of a much faster training time since these methods avoids any auxiliary optimization steps.
To further explore optimization, we found that optimization behaves similarly for all of the missing instances.
%The auxiliary optimization seeks to minimize (\ref{eqn:neutralOpt}), and to get a better sense of the optimization we keep track of the obtained norms throughout training.
Figure \ref{fig:l2norm} shows how the norm drops during training for 5 randomly chosen examples. These were taken from the Case 2 data, and selected from all sub-bags.

The overall structure of the network as shown in Figure \ref{fig:revnet1} provides a relatively simple path toward a model-parallel implementation. 
Our chosen implementation strategy places the execution of all operations pertaining to a network $C_i$ including the embedding aggregation, on its own GPU. 
Likewise, the concatenation of the aggregated embeddings and the fully connected layers $H$ are placed on their own GPU. 
Because of the dependency of $H$ on all the $C_i$ one could argue about reducing the idle resources by instead co-locating $H$ with one of the $C_i$.  However, that would have the downside of an imbalance in the model-based memory requirements.  
With $s=3$ our design requires a total of four GPUs. 
%The model parallelization over the GPUs was readily supported through the Keras library given that our hardware provided direct PCIe connectivity between groups of four GPUs. 
We also note that optimization for several missing images within the same bag similarly fits  this parallelization scheme.
Thus, the extra computational time required to solve these optimization problems is independent of the number of sub-bags and we observed approximately 40 seconds per bag.

As previously mentioned, there is a drop in accuracy after Phase 1 when switching from testing on full configurations ${\cal B}_T\cap {\cal B}_{\hat{c}}$ to testing on all other configurations, ${\cal B}_T \setminus {\cal B}_{\hat{c}}$.
Part of this drop is due to the model not having seen bags with missing sub-bags before, but another explanation is that the set ${\cal B}_T \setminus {\cal B}_{\hat{c}}$ has less complete information because of the missing sub-bag(s).
To test this explanation, we apply the dropout to each sub-bag individually on full-configurations and note the drop in accuracy.
Using Case 2 and the Shapenet dataset with the weights from the end of Phase 2, Table \ref{tab:dropout} shows the accuracy drops for each of the three sub-bags.
After training over all configurations, these results show that all of the sub-bags contribute to the accuracy over full configurations, and also that the sub-bags are not equally predictive.

\bibliography{nmi_2019}
\bibliographystyle{aaai}

\end{document}